\documentclass[letterpaper, 10 pt, conference]{ieeeconf}  
\usepackage{times}

\IEEEoverridecommandlockouts                              

\usepackage{multicol}
\usepackage[bookmarks=true]{hyperref}
\usepackage{graphicx}
\usepackage{amsfonts}
\usepackage{amsmath,amssymb} 
\usepackage{algorithm}
\usepackage[noend]{algpseudocode}
\usepackage{multirow}
\usepackage{caption}
\usepackage{subcaption}
\usepackage{soul} 
\usepackage{bchart}
\usepackage{newfloat}
\usepackage{bbm}
\usepackage{pgfplots}
\pgfplotsset{width=9cm,compat=1.8}

\DeclareFloatingEnvironment[
   fileext=loc
]{chart}

\usepackage[colorinlistoftodos,prependcaption,textsize=tiny]{todonotes}

\usepackage[normalem]{ulem}

\begin{document}

\title{Out-of-Distribution Robustness with Deep Recursive Filters}

\author{Kapil D. Katyal$^{1,2}$, I-Jeng Wang$^{1}$, and Gregory D. Hager$^{2}$ 
\thanks{$^{1}$Johns Hopkins University Applied Physics Lab, Laurel, MD, USA.
        {\tt\small Kapil.Katyal@jhuapl.edu}}%
\thanks{$^{2}$Dept. of Comp. Sci., Johns Hopkins University, Baltimore, MD, USA.
        }%
}

\maketitle

\begin{abstract}
Accurate state and uncertainty estimation is imperative for mobile robots and self driving vehicles to achieve safe navigation in pedestrian rich environments. A critical component of state and uncertainty estimation for robot navigation is to perform robustly under out-of-distribution noise. Traditional methods of state estimation decouple perception and state estimation making it difficult to operate on noisy, high dimensional data.  Here, we describe an approach that combines the expressiveness of deep neural networks with principled approaches to uncertainty estimation found in recursive filters.  We particularly focus on techniques that provide better robustness to out-of-distribution noise and demonstrate applicability of our approach on two scenarios: a simple noisy pendulum state estimation problem and real world pedestrian localization using the nuScenes dataset~\cite{nuscenes2019}. We show that our approach improves state and uncertainty estimation compared to baselines while achieving approximately $3\times$ improvement in computational efficiency. 

\end{abstract}

\section{Introduction}
State estimation is a critical problem impacting a wide variety of robotic applications including mapping, localization, pose estimation, and motion planning. These challenging applications are encumbered by issues such as high dimensional observations, partial observability, and noisy measurements. Traditional methods of state estimation, including Kalman and other recursive filters, decouple perception and state estimation by operating directly on low dimensional state representations after the perception pipeline.  This limits the ability to develop state estimation techniques that are robust to high sensor noise.

Deep learning techniques have made a significant contribution in many areas including perception, speech recognition, and robotics. Particularly, they offer the ability to operate directly on high dimensional spaces where each layer represents learned features that can be used to perform functions such as classification and regression.  While deep learning techniques have demonstrated significant improvements in predictive accuracy, they have demonstrably fallen short in estimating the predictive uncertainty. As demonstrated by Ovadia and Fertig~\cite{NIPS2019_9547}, this is particularly exacerbated when evaluating predictive uncertainty under distributional shift. For many applications, including robotics, this is a fundamental limitation as robustness to out-of-distribution noise is critical to improve the safety and reliability of systems when deployed into the real world.

In this paper, we study methods that combine deep neural networks with traditional state estimation techniques that improve robustness to noisy input data.  We specifically focus on developing techniques that capture aleatoric (uncertainty found in observations) and epistemic uncertainty (model uncertainty)~\cite{NIPS2017_7141} to improve prediction interval accuracy in the presence of noisy, out-of-distribution inputs.  

Our specific focus is along three core research questions:

\begin{enumerate}

 \item Can principles of recursive filters combined with neural networks improve state estimation accuracy of dynamical systems in the presence of out of distribution noise?
 \item Can this approach improve the robustness and performance of predictive uncertainty estimation as the test distribution deviates from the training distribution?
 \item Can properties of the recursive filter provide additional insight regarding expected competency of the trained network?

\end{enumerate}
\begin{figure}[t] 
	\centering
    \includegraphics[width=0.8\columnwidth]{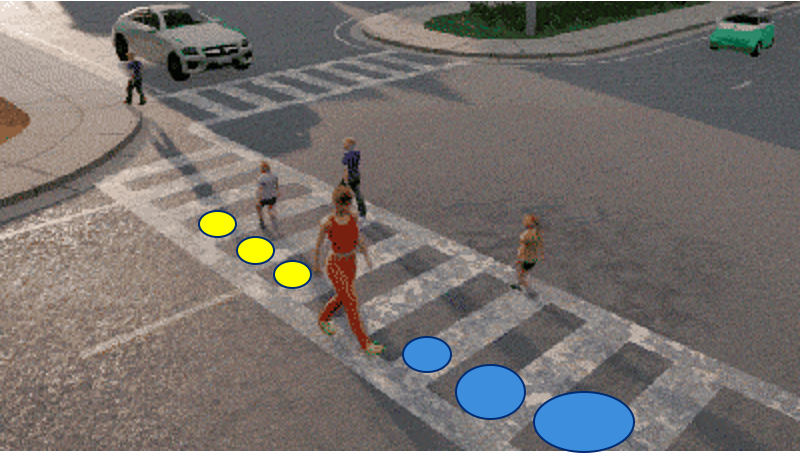}
    \caption{The goal of this project is to develop robust state and uncertainty estimation for pedestrian localization by combining elements of deep neural networks, recursive filters and uncertainty estimation.}
	\label{fig:teaser}
\end{figure}

\begin{figure*}[t] 
	\centering
	    \hspace{10mm}
    \includegraphics[width=1.9\columnwidth]{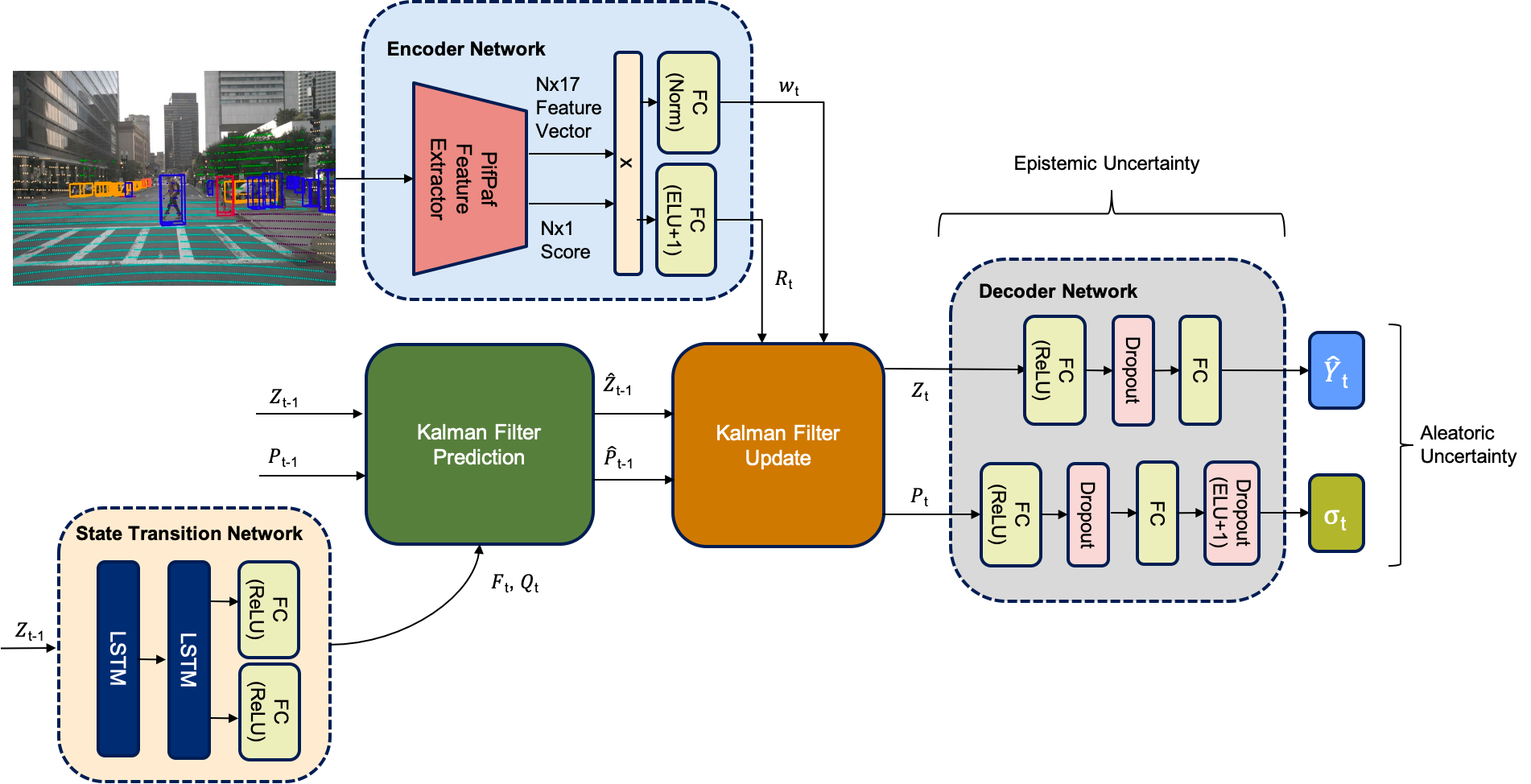}
    \caption{This figure describes our overall network architecture for the pedestrian localization experiment.  It consists of an encoder network used to learn a latent embedding and measurement covariance matrix, a state transition network used to learn a dynamics model and process covariance matrix and a decoder network used to decode the latent embedding to estimated depth along with capturing aleatoric and epistemic uncertainty.  }
	\label{fig:architecture}
\end{figure*}

To study these questions, we evaluate our approach using two experimental environments. The first is a toy problem using a simulated pendulum.  The objective is to estimate the pendulum angle in the presence of varying observation noise added to the  $24\times24$ pixel images as input. Our goal is to assess the effects of a learned dynamical model and demonstrate robustness during regions of high observation noise.  

We then focus on a real world problem by estimating the 3D pedestrian localization from monocular images.  For this evaluation, we use the nuScenes~\cite{nuscenes2019} dataset consisting of real world driving scenarios with corresponding annotations including pedestrian position. Again, we assess the ability of our model to provide robust state and uncertainty estimation in the presence of out-of-distribution noise for this more complex, real world scenario.  

The following summarizes our main contributions:
\begin{enumerate}

\item We develop an architecture, referred to as deep recursive filter (DRF) that combines deep neural networks to learn latent embeddings, state transition models, and associated covariance matrices with a recursive filter operating on the learned latent states.
\item Our approach explicitly models aleatoric and epistemic uncertainty as part of the deep recursive filter to improve confidence interval prediction.
\item We show through experimental evidence that our combined approach improves state estimation, confidence interval estimation, and runtime performance on a toy problem and real world pedestrian localization.
\item Further, we show that our approach is more robust to out-of-distribution noise compared to state-of-the-art approaches.

\end{enumerate}

\section{Related Work}

\subsection*{State Space Modeling}
Classical state estimation has been instrumental in solving a wide variety of problems in the robotics community and can be summarized by several review papers including~\cite{thrun2005probabilistic,kalmanfilterreview,5985520}. Recently, many research efforts introduced concepts that combine neural networks with state space models and recursive filters. BackpropKF~\cite{NIPS2016_6090} uses a feedforward neural network to produce a latent embedding and covariance matrix from a raw high dimensional input however uses a known state transition model. This is further extended by DPF~\cite{jonschkowski18} which includes the use of particle filters for state estimation with a known dynamics model. Additional works combine variational autoencoders with Kalman Filters including~\cite{krishnan2015deep,NIPS2017_6951}. The work presented in~\cite{DBLP:journals/corr/abs-1905-07357} develop a Recurrent Kalman Filter Framework that learns a transition model operating on latent representations with an emphasis factorized inference for efficient computation. DVBF is another work~\cite{dvbf_inproceedings} that learns state space models using Bayesian Filters for stable long term predictions. Recently introduced, DynaNet~\cite{chen2020dynanet} develop state estimation and motion prediction techniques using a neural Kalman Filter and evaluate their approach on visual odometry using the KITTI dataset.

\subsection*{Uncertainty Estimation}

Bayesian Neural Networks~\cite{richard1991neural,10.5555/525544}  have been used extensively to represent uncertainty in a neural network by learning a probability distribution over the network parameters. Recent works also focus on improving uncertainty estimation using deep neural networks. Kendall and Gal~\cite{NIPS2017_7141} describe techniques that capture aleatoric and epistemic uncertainty particularly in the computer vision domain. Gal and Ghahramani~\cite{pmlr-v48-gal16} describe uncertainty estimation techniques using stochastic dropout as a form of Bayesian approximation. Ovadia and Fertig~\cite{NIPS2019_9547} evaluate the performance of predictive uncertainty under distributional shift on a variety of modalities including images, text and categorical data.

\subsection*{Monocular 3D State Estimation}

Several techniques focus on performing state estimation with a monocular camera. Engel et al.~\cite{engel14eccv} develop large scale SLAM algorithms that operate on a monocular camera with an emphasis on real-time performance. ORB-SLAM~\cite{7219438} extends this work to focus on robustness to motion clutter using loop closure and relocalization. Bertoni et al. develop the Monoloco algorithm~\cite{Bertoni_2019_ICCV} which is among the first approaches to perform 3D pedestrian localization from monocular images that captures both aleatoric and epistemic uncertainty, however do so without modeling dynamics.


Our approach is unique from other works in that we include explicit modeling of aleatoric and epistemic uncertainty within the deep recursive filter framework.  We show this combined uncertainty modeling with learned dynamics from a recursive filter framework is imperative to improve state estimation and prediction interval robustness in the presence of out-of-distribution noise. Further, we show that these techniques apply not only to toy problems but also the challenging problem of pedestrian localization using monocular cameras.  Finally, we investigate whether properties of a recursive filter can be used as a measure of competency to further improve computational efficiency of the uncertainty estimation.

\section{Preliminaries}

Our objective is to perform state and uncertainty estimation operating directly on high-dimensional observations.  Formally, we define the noisy pendulum problem as follows. At given time $t$, we observe a $24\times24$ pixel image with noise added to the image.  Our objective is to estimate the state of the pendulum defined as: $X_{t} = (cos(\theta_{t}), sin(\theta_{t}))^T$ as well as the predicted uncertainty of each state variable, $\sigma_{cos, t}$ and $\sigma_{sin, t}$. 

For the pedestrian localization experiments, we define the state of the pedestrian $i$ at time $t$ as $X_{i,t} = (x_{i,t}, y_{i,t}, z_{i,t})$, the Cartesian position of the pedestrian. Similar to~\cite{Bertoni_2019_ICCV}, we assume a calibrated camera provides the transformation from image coordinates to the $x$ and $y$ Cartesian points.  The trained neural network receives a $1600 \times 900$ 3-channel RGB image and estimates the $z_{t}$ coordinate corresponding to the depth of the pedestrian using a single monocular image as well as the aleatoric and epistemic uncertainty of the depth estimate, $\sigma_{depth, t}$.  

\begin{figure}
    \centering
    \hspace{7mm}
    \includegraphics[width=0.8\columnwidth]{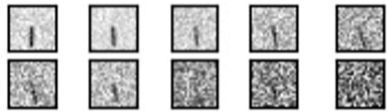}
    \vspace{0.7cm}

\begin{tikzpicture}
\begin{axis}[
    ybar=0pt,
    bar width=10pt,
    width=6cm,height=5cm,
    enlarge x limits=0.6,
    enlarge y limits=0.05,
    legend style={at={(0.5,-0.20)},
      anchor=north,legend columns=-1},
    ylabel={MAE},
    xlabel={Noise},
    symbolic x coords={0.5,0.75},
    xtick=data,
    nodes near coords align={vertical},
     every axis plot/.append style={
          fill}
    ]
\addplot[red!20!black,fill=red!60!white] coordinates {(0.5,0.04651897668)(0.75,0.100267044)};
\addplot[green!20!black,fill=green!60!white] coordinates {(0.5,0.03282828297)(0.75,0.06742980505)};
\addplot[blue!20!black,fill=blue!60!white] coordinates {(0.5,0.004554806631)(0.75,0.0426968253)};
\legend{No Dynamics,LSTM,Ours}
\end{axis}
\end{tikzpicture}  

       

\begin{tikzpicture}
\begin{axis}[
    xlabel={MPIW},
    ylabel={PICP},
    xmin=0, xmax=5,
    ymin=90, ymax=100,
    width=7cm,height=5cm,
    xtick={1,2,3,4},
    ytick={90, 95, 100},
    legend pos=south east,
    ymajorgrids=true,
    grid style=dashed,
]

\addplot[
    color=red,
    mark=square,
    ]
    coordinates {
    (1.63,96.6)(2.48,98.3)(3.34,99.1)
    };
    \addlegendentry{No Dynamics}

\addplot[
    color=green,
    mark=square,
    ]
    coordinates {
    (1.70,94.5)(2.46,98.4)(3.27,99.2)
    };
    \addlegendentry{LSTM}

\addplot[
    color=blue,
    mark=square,
    ]
    coordinates {
    (1.44,97.5)(2.18,98.7)(2.87,99.2)
    };
    \addlegendentry{Ours}

\end{axis}
\end{tikzpicture}
\caption{(Top) Samples of the noisy pendulum. (Middle) The figure compares the MAE between the DRF approach with the baselines architectures for the noisy pendulum problem for the out-of-distribution noise experiments.  (Bottom) This figure compares the PICP metric with the MPIW metric for the three architectures.  We show our approach significantly reduces state estimation error and improves to the uncertainty estimation compared to the baselines.}
\label{table:mae_chart_pend}
\end{figure}
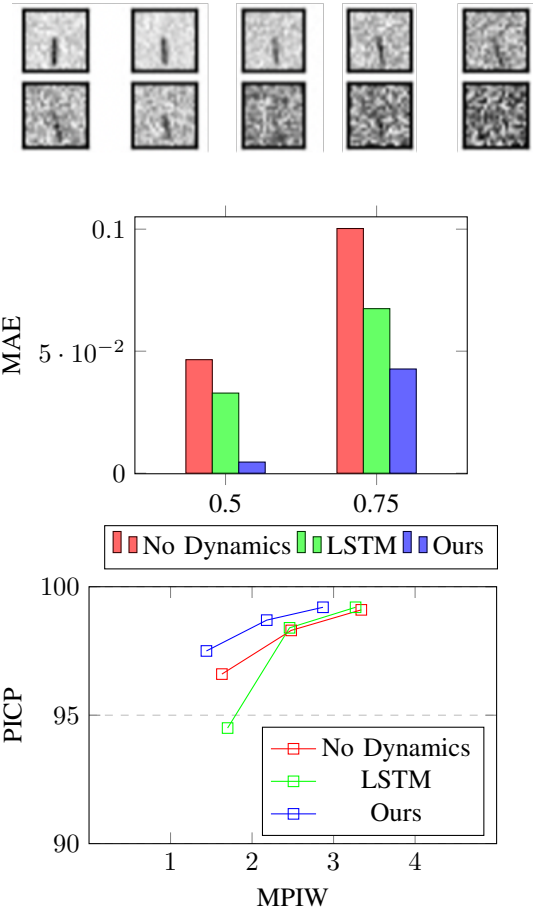

\subsection{Metrics}
Our objective is to improve state and uncertainty estimation while developing techniques that are computationally efficient.  To evaluate the accuracy of state estimation, we measure the Mean Absolute Error (MAE) (Eq.~\ref{eq:mae}) between the predicted and ground truth states. We use the average execution time to compare the computational efficiency of various approaches. To assess the quality of uncertainty estimation, we use the Prediction Interval Coverage Probability (PICP)~\cite{pmlr-v80-pearce18a,khosravi2010prediction} and Mean Prediction Interval Width (MPIW)~\cite{pmlr-v80-pearce18a,DBLP:journals/corr/abs-1806-11222} metrics as defined by Eq.~\ref{eq:picp} and ~\ref{eq:mpiw}. Here, $l(x_i)$ and $u(x_i)$  are the lower and upper bounds of the confidence interval respectively, $\hat{y}_{i}$ is the point state estimate, $y_i$ is the ground truth, and $\mathbbm{1}$ is the indicator function.  

\begin{equation} \label{eq:mae}
\mathbf{MAE} := \frac{1}{N}\sum_{i=1}^{N}\left | \hat{y}_{i} - y_{i} \right |
\end{equation}

\[
c_i :=  \mathbbm{1}({l(x_i) \leq y_i \leq u(x_i) })
\]
\begin{equation} \label{eq:picp}
\mathbf{PICP}_{{l(x)}, {u(x)}}  :=  
		\frac{1}{N} \sum_{i=1}^{N}{c_i}\text{,}\end{equation}

\begin{equation} \label{eq:mpiw}
\mathbf{MPIW}_{{l(x)}, {u(x)}}  :=  
		\frac{1}{N} \sum_{i=1}^{N}{|u(x_i) - l(x_i)|}\end{equation}

Intuitively, the PICP metric measures the percentage of samples where the ground truth falls inside the predicted confidence interval and the MPIW metric measures the average width of the predicted interval. The goal is to maximize PICP while minimizing the MPIW metric.

\section{Approach}\label{approach}

Our algorithm architecture, inspired by~\cite{chen2020dynanet} is described in Fig.~\ref{fig:architecture} which consists of encoder, decoder and state transition neural networks along with the Kalman Filter (KF) prediction and update steps.  

\subsection*{Encoder Network}

The encoder network receives a $24 \times 24 \times 1$ grayscale image in the pendulum experiments or a $1600 \times 900 \times 3$ RGB image for the pedestrian localization experiments to produce a latent embedding representing and an associated observation noise covariance matrix. In the pendulum experiment, the encoder network consists of a $5\times5$ convolutional layer and a $3\times3$ convolutional layer.  This is followed by a fully connected layer to produce a feature vector $\mathbf{x\_in}_{i,t}$.  For the pedestrian localization experiments, similar to~\cite{Bertoni_2019_ICCV}, we first extract PifPaf~\cite{kreiss2019pifpaf} features that represent a $17 \times 2$ dimensional keypoint vector ($x_{i,t}$, $y_{i,t}$) for each pedestrian $i$ along with a confidence score. They keypoint vector and confidence scores are concatenated to produce feature vector $\mathbf{x\_in}_{i,t}$.  In both experiments, we pass the feature vector $\mathbf{x\_in}_{i,t}$ to two fully connected layers to generate a latent embedding $\mathbf{w}_{i,t}$ for each pedestrian along with a covariance matrix $\mathbf{R}_{i,t}$ that represents the observation noise.  The output features of $Linear_1$ and $Linear_2$ are both 30 dimensions.

\[\mathbf{w}_i^t = {Layer\_Norm}({Linear_1}(\mathbf{x\_in}_{i,t}))\]
\[\mathbf{R}_i^t = Diag[{Elu}({Linear_2}(\mathbf{x\_in}_{i,t})) + 1]\]

\subsection*{State Transition Network}

The main goal of the state transition network is to learn a dynamical model in the latent embedding,  $\mathbf{w}_{i,t}$,  through a  learned state transition matrix, $\mathbf{F}_t$ and associated process noise covariance matrix $\mathbf{Q}_t$. We accomplish this by using two LSTM layers followed by a fully connected layer for each output, $\mathbf{F}_t$ and $\mathbf{Q}_t$.

\subsection*{Kalman Filter Prediction and Update}

The Kalman filter prediction step follows the principles of traditional Kalman filters where the objective is to produce an a priori estimate of the state and estimate covariance matrix, $\mathbf{\hat{z}}_{t-1}$ and $\mathbf{\hat{P}}_{t-1}$ respectively using the learned state transition matrix.  The state transition matrix and associated process noise covariance matrices, $\mathbf{F}_t$ and $\mathbf{Q}_t$ are learned by the state transition network. The following operations are performed on the previous latent embedding, $\mathbf{z}_{t-1}$.

\[\mathbf{\hat{z}}_{t-1} = \mathbf{\hat{F}}_{t}\mathbf{{z}}_{t-1}\]
\[\mathbf{\hat{P}}_{t-1} = \mathbf{\hat{F}}_{t}\mathbf{{P}}_{t-1}\mathbf{\hat{F}}_{t}^T + \mathbf{Q}_t\]

The KF update step produces a posteriori estimate of the latent embedding and estimate covariance matrix, ${\mathbf{z}}_{t-1}$ and $\mathbf{P}_{t-1}$ according to the following equations.

\[\mathbf{\tilde{y}}_t = \mathbf{w}_t - \mathbf{H}\mathbf{z}_{t-1}\]
\[\mathbf{S}_t = \mathbf{H}_t\mathbf{\hat{P}}_{t-1}\mathbf{H}_t^T+\mathbf{R}_t\]
\[\mathbf{K}_t = \mathbf{\hat{P}}_{t-1}\mathbf{H}_t^T\mathbf{S}_t^{-1}\]
\[\mathbf{z}_t = \mathbf{z}_{t-1} + \mathbf{K}_t\mathbf{\tilde{y}}_t\]
\[\mathbf{P}_t = (\mathbf{I} - \mathbf{K}_t\mathbf{H}_t)\mathbf{\hat{P}}_{t-1}\]

\begin{figure}

\begin{tikzpicture}
\begin{axis}[
    ybar,
    bar width=15pt,
     enlarge x limits=0.25,
    enlarge y limits=0.05,
    legend style={at={(0.5,-0.20)},
      anchor=north,legend columns=-1},
    ylabel={MAE},
    xlabel={Noise},
    symbolic x coords={0.01,0.025,0.05},
    xtick=data,
    nodes near coords align={vertical},
    ]
\addplot[red!20!black,fill=red!60!white] coordinates {(0.01,3.42) (0.025,2.67) (0.05,3.39)};
\addplot[green!20!black,fill=green!60!white] coordinates {(0.01,2.88) (0.025,2.44) (0.05,2.98)};
\addplot[orange!20!black,fill=orange!60!white] coordinates {(0.01,2.56) (0.025,2.32) (0.05,2.87)};
\addplot[blue!20!black,fill=blue!60!white] coordinates {(0.01,2.56) (0.025,2.26) (0.05,2.72)};
\legend{No Dynamics,LSTM,Monoloco, Ours}
\end{axis}
\end{tikzpicture}
\caption{Comparison of MAE for pedestrian localization with a changing distribution of noise. In these experiments, our DRF approach performs comparably to Monoloco~\cite{Bertoni_2019_ICCV} for in-distribution evaluation and outperforms all baselines for out-of-distribution evaluation.}
  \label{fig:mae_chart_nuscenes}
\end{figure}
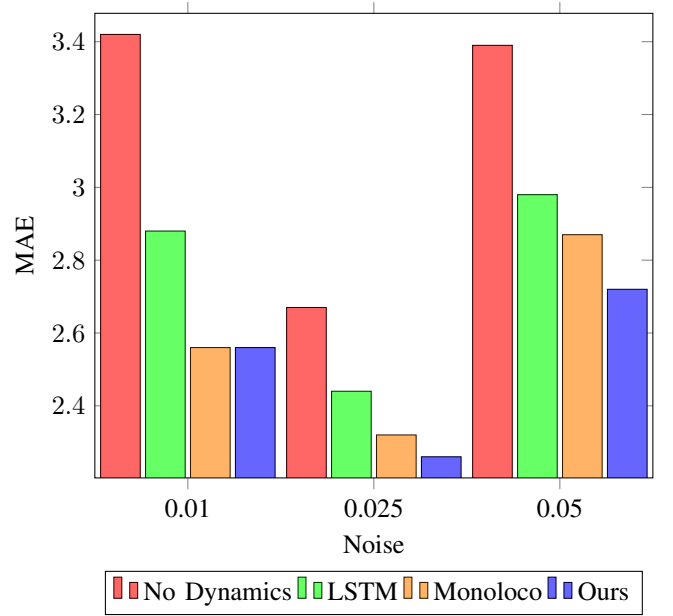

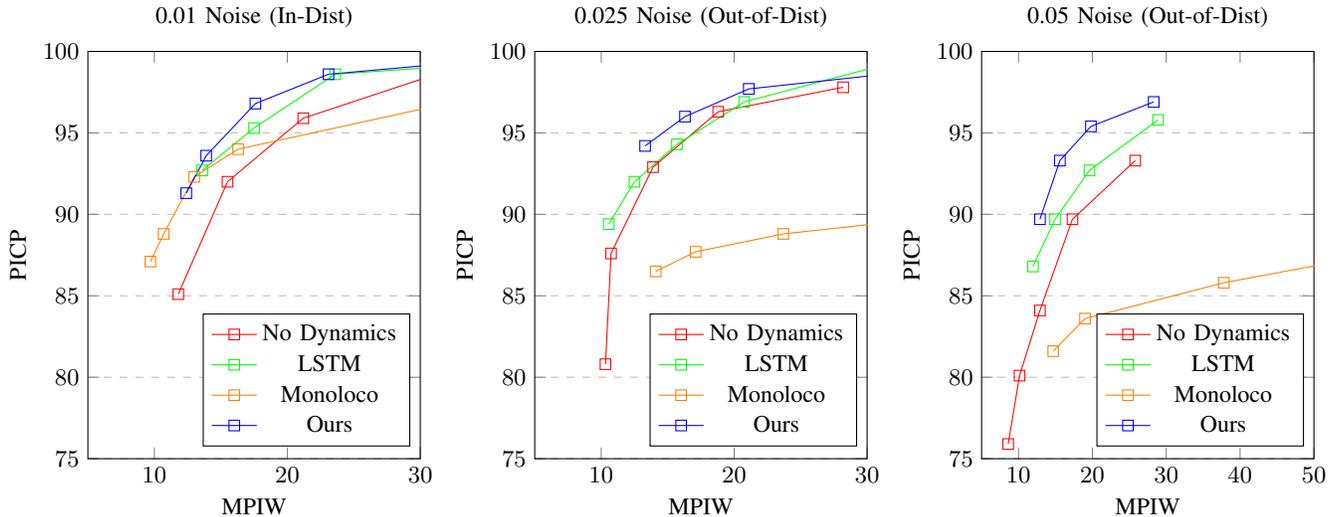
\begin{figure*}\small
\centering

\begin{tikzpicture}
\begin{axis}[
    title={0.01 Noise (In-Dist)},
    xlabel={MPIW},
    ylabel={PICP},
    xmin=5, xmax=30,
    ymin=75, ymax=100,
    width=6cm,height=7cm,
    xtick={10,20,30,40,50},
    ytick={75, 80,85, 90, 95, 100},
    legend pos=south east,
    ymajorgrids=true,
    grid style=dashed,
]
    \legend{};

\addplot[
    color=red,
    mark=square,
    ]
    coordinates {
    (32.3,98.9)(21.2,95.9)(15.5,92)(11.8,85.1)
    };
    \addlegendentry{No Dynamics}

\addplot[
    color=green,
    mark=square,
    ]
    coordinates {
    (13.6,92.7)(17.5,95.3)(23.6,98.6)(35.8,99.3)
    };
    \addlegendentry{LSTM}

\addplot[
    color=orange,
    mark=square,
    ]
    coordinates {
    (9.7, 87.1)(10.7,88.8)(13,92.3)(16.3, 94)(32, 96.8)
    };
    \addlegendentry{Monoloco}
    
\addplot[
    color=blue,
    mark=square,
    ]
    coordinates {
    (12.4, 91.3)(13.9, 93.6)(17.6,96.8)(23.1, 98.6)(34.0, 99.4)
    };
    \addlegendentry{Ours}

\end{axis}
\end{tikzpicture}
\begin{tikzpicture}
\begin{axis}[
    title={0.025 Noise (Out-of-Dist)},
    xlabel={MPIW},
    ylabel={PICP},
    xmin=5, xmax=30,
    ymin=75, ymax=100,
    width=6cm,height=7cm,
    xtick={10,20,30,40,50},
    ytick={75, 80,85, 90, 95, 100},
    legend pos=south east,
    ymajorgrids=true,
    grid style=dashed,
]

\addplot[
    color=red,
    mark=square,
    ]
    coordinates {
    (28.2,97.8)(18.8,96.3)(13.9,92.9)(10.74,87.6)(10.32,80.8)
    };
    \addlegendentry{No Dynamics}

\addplot[
    color=green,
    mark=square,
    ]
    coordinates {
    (30.9,99.1)(20.75,96.9)(15.7,94.3)(12.5,92.0)(10.57,89.4)
    };
    \addlegendentry{LSTM}

\addplot[
    color=orange,
    mark=square,
    ]
    coordinates {
    (14.1,86.5)(17.1,87.7)(23.7,88.8)(40.6,90.3)
    };
    \addlegendentry{Monoloco}
    
\addplot[
    color=blue,
    mark=square,
    ]
    coordinates {
    (30.2, 98.5)(21.1, 97.7)(16.3,96)(13.3,94.2)
    };
    \addlegendentry{Ours}

\end{axis}
\end{tikzpicture}
\begin{tikzpicture}
\begin{axis}[
    title={0.05 Noise (Out-of-Dist)},
    xlabel={MPIW},
    ylabel={PICP},
    xmin=5, xmax=50,
    ymin=75, ymax=100,
    width=6cm,height=7cm,
    xtick={10,20,30,40,50},
    ytick={75, 80,85, 90, 95, 100},
    legend pos=south east,
    ymajorgrids=true,
    grid style=dashed,
]

\addplot[
    color=red,
    mark=square,
    ]
    coordinates {
    (8.6, 75.9)(10.14,80.1)(12.9,84.1)(17.3,89.7)(25.8,93.3)
    };
    \addlegendentry{No Dynamics}

\addplot[
    color=green,
    mark=square,
    ]
    coordinates {
    (11.95,86.8)(14.95,89.7)(19.6, 92.7)(28.9,95.8)
    };
    \addlegendentry{LSTM}

\addplot[
    color=orange,
    mark=square,
    ]
    coordinates {
    (14.7, 81.6)(19,83.6)(37.8,85.8)(99.1, 91)
    };
    \addlegendentry{Monoloco}
    
\addplot[
    color=blue,
    mark=square,
    ]
    coordinates {
    (12.9, 89.7)(15.6, 93.3)(19.8,95.4)(28.3, 96.9)
    };
    \addlegendentry{Ours}

\end{axis}
\end{tikzpicture}
\caption{A comparison of the PCIP versus MPIW metrics for each algorithm for in and out-of-distribution noise. The data points are generated by varying the dropout rate when computing the epistemic uncertainty.  All three algorithms generally show comparable performance for in-distribution samples, however DRF performs better as the noise distribution increases. }
  \label{fig:roc_curves}
\end{figure*}

\subsection*{Decoder Network}

The decoder network has two purposes: (1) receive the posteriori estimate of the latent embedding and covariance matrix to decode an estimated state, (2) to calculate the associated aleatoric and epistemic uncertainties of the state estimate. Aleatoric uncertainty is computed using the mean variance estimation (MVE)~\cite{374138} technique where the decoder produces two outputs that represent the mean and variance of a Normal distribution that are sampled during inference.  To calculate the epistemic uncertainty, we use a stochastic dropout technique~\cite{pmlr-v48-gal16} where several dropout layers were added to the decoder. During inference, multiple passes of the neural network generate a sample population from which the mean and variance can be calculated. The decoder network can be described as follows:

\[\mathbf{h}_{i,t} = {ReLU}({Linear_3}([\mathbf{z}_{i,t},\mathbf{P}_{i,t}]))\]
\[\mathbf{y}_{i,t} = {Linear_4}(\mathbf{h}_{i,t})\]
\[\mathbf{\sigma}_{i,t} = {Elu}({Linear_5}(\mathbf{h}_{i,t})) + 1\]

\subsection*{Training Details}

The entire network is trained using the Pytorch~\cite{NEURIPS2019_9015} framework with the Adam optimizer~\cite{DBLP:journals/corr/KingmaB14} with the learning rate set to $10^{-4}$ for 200 epochs.   We use the negative log likelihood (NLL) of a Normal distribution as the loss function computed at the end of a sequence of length $S$ according to Eq.~\ref{eq:loss_function}.

\begin{equation} \label{eq:loss_function}
\mathcal{L} = \frac{1}{S}\sum_1^S{(log(\sqrt{2\pi\sigma^2_s}) + \frac{1}{2\sigma^2_s}\sum{(y_s-\hat{y_s})^2})}
\end{equation}

\begin{figure*}[t] 
	\centering
    \includegraphics[width=0.73\columnwidth]{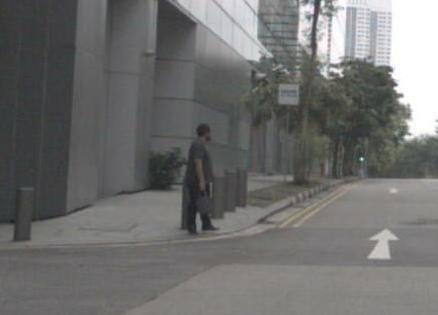}
    \includegraphics[width=0.73\columnwidth]{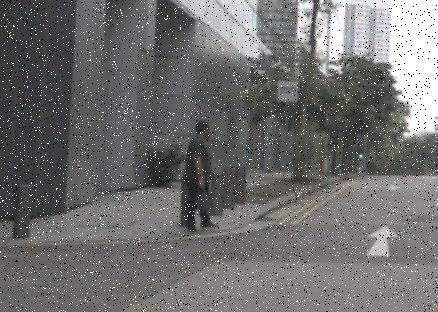}
    \caption{(Left) Representative image from nuScenes dataset~\cite{nuscenes2019}. (Right) Distorted image with noise added.}
	\label{fig:noisy_img}
\end{figure*}

\section{Noisy Pendulum Experimental Evaluation}
We first evaluate out-of-distribution robustness on the noisy pendulum problem similar to ~\cite{DBLP:journals/corr/abs-1905-07357,NIPS2017_6951}. For the training set, we generate 2000 sequences of length 75 corresponding to pendulum motion. The dataset includes a $24 \times 24 \times 1$ image as the input and the corresponding state of the pendulum as defined by $X_{t} = (cos(\theta_{t}), sin(\theta_{t}))^T$. For each sequence, random Gaussian noise was added to the images.  We follow the noise generating procedure described in~\cite{DBLP:journals/corr/abs-1905-07357} where we vary the maximum amount of noise between 0 and 50\% during training.  

To measure robustness to out-of-distribution noise, during evaluation, we set the maximum noise threshold to 75\% to simulate out-of-distribution noise.  We compare against two baselines. The first baseline, referred to as no\_dynamics, removes any aspects of a learned dynamical model. Here we map the outputs of the encoder network, $\mathbf{w}_t$ and $\mathbf{R}_t$ directly to the decoder network and train end-to-end in a similar fashion.  Our second baseline replaces the Kalman filter elements (e.g., State Transition Network, Kalman Filter Prediction, and Kalman Filter Update steps) with a vanilla LSTM.  As demonstrated in Fig.~\ref{table:mae_chart_pend}, we show significant improvement to state estimation in both the in and out-of-distribution noise profile.  Further, the recursive filter demonstrates better robustness to predicting confidence intervals.

\section{Pedestrian Localization Experimental Evaluation}

We pivot our evaluation to focus on pedestrian localization. For this experiment, we use the nuScenes  (part 1) training and validation dataset~\cite{nuscenes2019} with tracked pedestrians per key frame. This consists of approximately 3300 images with approximately 3700 pedestrian instances.  We add Gaussian noise to these images to represent out-of-distribution evaluation.  We trained with a probability of 0.01 that a pixel will be replaced with either a black or white pixel. We then evaluated performance with probabilities of 0.025 and 0.05 to represent out-of-distribution noise (example found in Fig.~\ref{fig:noisy_img})\footnote{We experimented with higher probabilities of noise, however this caused PifPaf detector to miss a significant amount of detections. Future work will bypass this feature extractor layer.}.

\begin{table}
  \centering
   \setlength{\tabcolsep}{1.2mm}
   \renewcommand{\arraystretch}{1.1}
  \begin{tabular}{|c|c|c|c|}
  \hline
  \textbf{Algorithm} &  \textbf{Average Time (ms) $\downarrow$} & \textbf{Speed Up}\\
  
  \hline
  \hline

 Monoloco  & 36.1 & 1x \\
  \hline
  Ours & \textbf{12.3} & \textbf{2.93x} \\
  
  \hline

   \end{tabular}
  \caption{The average execution time for 25 forward passes. These numbers are computed after the PifPaf feature extraction which is common to both approaches and measured using an NVIDIA\textsuperscript{\tiny\textregistered} GTX 1080 GPU.}
  \label{table:performance}
\end{table}

\subsection{Baseline Comparisons} 

We compare our algorithm to several baselines for comparison purposes. The first two baselines, no\_dynamics and LSTM are similar to the baselines used in the noisy pendulum experiments. The third baseline that we use for comparison is Monoloco~\cite{Bertoni_2019_ICCV} which similarly estimates pedestrian localization with uncertainty.

\subsection{State Estimation} 

We first measure the MAE of the estimated depth of the pedestrian compared to the ground truth provided by the nuScenes dataset.  In Fig.~\ref{fig:mae_chart_nuscenes}, we compare our approach to the {no\_dynamics}, {LSTM} and the Monoloco work. When evaluating in distribution, our approach is comparable to Monoloco, with both approaches outperforming the {no\_dynamics} and {LSTM} baselines.  When evaluating out-of-distribution, we show an 18\% improvement in the MAE compared to the no\_dynamics baseline, 8\% improvement compared to the LSTM baseline, and a 4\% improvement compared to Monoloco.

\subsection{Uncertainty Estimation} 

As described in Sec. \ref{approach}, we use the PICP and MPIW metrics to assess the quality of the prediction intervals. These results are summarized in Fig. ~\ref{fig:roc_curves}, where we analyze the trade-offs between PICP and MPIW metrics for each algorithm across in and out-of-distribution noise.  For the in-distribution case, all four algorithms exhibit similar performance in uncertainty estimation.  When evaluating out-of-distribution, the performance begins to diverge with our approach showing an distinct improvement to the other methods in particularly high out-of-distribution noise samples. 

\subsection{Computational Efficiency}

An important criteria is to allow uncertainty estimation to run in real time.  For that purpose, we also evaluate the run time performance of our approach as summarized by Table~\ref{table:performance}.  We demonstrate an approximately 3x improvement largely due to the simpler decoder network used to perform epistemic uncertainty.



\section{Discussion}

Accurate and robust confidence interval prediction for state estimation in presence of out-of-distribution noise is critical to the deployment of robotic systems into the real world.  In this paper, we show that by modeling the recursive filter framework within a neural network architecture and capturing aleatoric and epistemic uncertainty, we can significantly improve state estimation and the robustness of uncertainty estimation.  In the pedestrian localization scenario, we show an average of 4\% improvement to the MAE across the varying noise distributions compared to next highest performing baseline and 18\% improvement compared to the no\_dynamics baseline. Further, in the presence of distributional shift, we show higher percentage of samples found in the prediction interval while also reducing the overall prediction interval width.  Finally, we show our technique is more computationally efficient as it computes aleatoric and epistemic uncertainty with approximately 3x performance improvement compared to the baseline.

We believe an interesting area of future work remains in better understanding the learned covariance matrices and leveraging the Kalman gain as a measure of competency to improve uncertainty sampling performance.  The intuition is that one would only compute the expensive epistemic uncertainty when the estimated competency is low. Applying this technique on the pendulum experiments, we achieve good results.  As shown in Table~\ref{table:pendulum_kalman_gain}, the PCIP metric is comparable to always performing stochastic dropout, while improving the MPIW by 50\% and the runtime performance by 40\%. However, in the pedestrian localization experiment, we do not see the same performance. We believe this can be attributed to the sources of uncertainty. In the pendulum example, much of the observation uncertainty is represented by image noise and is easily captured by the Kalman gain. In the pedestrian example, there are far more sources of uncertainty beyond observation noise including distance from the camera and occlusions that are difficult to capture in the Kalman gain trained in an end-to-end manner. Our planned future work is to better characterize these uncertainties and use regularization techniques to further improve competency estimation with the goal of achieving similar performance as the pendulum example. Another opportunity for future work is to better leverage dynamical models. Here, we believe driving the learned state transition network towards a known dynamics model will improve robustness and sample efficiency and is an area we are continuing to further explore.

\begin{table}
  \centering
   \setlength{\tabcolsep}{1.2mm}
   \renewcommand{\arraystretch}{1.1}
  \begin{tabular}{|c|c|c|c|}
  \hline
  \textbf{Algorithm} &  \textbf{PICP $\uparrow$} & \textbf{MPIW $\downarrow$} & \textbf{Avg. Time $\downarrow$}\\
  
  \hline
  \hline
  
  Pend - (no dropout)  & 79.3 & 0.04 & 2.48 \\
  \hline
  Pend - (random)  & 86.3 & 0.30 & 17.1 \\
  
  \hline
  Pend - (thresh. by Kalman gain)  & 92.1 & 0.24 & 16.8 \\
  \hline
  
Pend - (always dropout) & 92.9 & 0.48 & 28.23 \\
  \hline
    \hline
     Ped - (no dropout)  & 86.0 & 10.95 & 1.63 \\
  \hline
  Ped - (random)  & 87.4 & 11.5 & 4.25 \\
  
  \hline
  Ped - (thresholded by Kalman gain)  & 87.7 & 11.6 & 4.06 \\
  \hline
  
Ped - (always dropout) & 89.7 & 12.9 & 11.3 \\
  \hline

   \end{tabular}
  \caption{This table captures the effects of using the Kalman gain as a measure of competency on choosing when to compute the epistemic uncertainty. }
  \label{table:pendulum_kalman_gain}
\end{table}

\section*{Acknowledgments}
This work was funded by the JHU Institute for Assured Autonomy and the JHU/APL Independent Research and Development Program.
\clearpage
\bibliographystyle{IEEEtran}
\bibliography{main}
\end{document}